\documentclass[10pt, a4paper, conference, compsocconf]{IEEEtran}

\usepackage{graphicx}
\usepackage{amsmath,amssymb,booktabs,tabularx}
\usepackage[hyphens]{url}
\usepackage{hyperref}
\usepackage{mathrsfs}
\usepackage{float}
\usepackage{multirow}
\usepackage{soul}
\usepackage[table]{xcolor}
\usepackage{comment}
\usepackage{subcaption}
\usepackage{mwe}
\usepackage{slashbox}
\usepackage{multicol}
\usepackage{import}
\usepackage{siunitx}
\usepackage{mathtools,url}
\usepackage{amssymb}

\usepackage{makecell} 

\begin{document}
%
\title{Hyper-Hue and EMAP on Hyperspectral Images for Supervised Layer Decomposition of Old Master Drawings}




%
\author{\IEEEauthorblockN{AmirAbbas Davari\IEEEauthorrefmark{1},
Nikolaos Sakaltras\IEEEauthorrefmark{1},
Armin Haeberle\IEEEauthorrefmark{2}, 
Sulaiman Vesal\IEEEauthorrefmark{1},
Vincent Christlein\IEEEauthorrefmark{1} \\
Andreas Maier\IEEEauthorrefmark{1} and
Christian Riess\IEEEauthorrefmark{1}}
\IEEEauthorblockA{\IEEEauthorrefmark{1}Pattern Recognition Lab, Department of Computer Science
	Friedrich-Alexander University \\ Erlangen Nuremberg,
	Erlangen, Germany, Email: amir.davari@fau.de}
\IEEEauthorblockA{\IEEEauthorrefmark{2}Bibliotheca Hertziana - Max-Planck-Institute for Art History, Rome, Italy}
}

\onecolumn
Copyright 2018 IEEE. Published in the 2018 International Conference on Image Processing (ICIP 2018), scheduled for October 7-10, 2018 in Athens, Greece. Personal use of this material is permitted. However, permission to reprint/republish this material for advertising or promotional purposes or for creating new collective works for resale or redistribution to servers or lists, or to reuse any copyrighted component of this work in other works, must be obtained from the IEEE. Contact: Manager, Copyrights and Permissions / IEEE Service Center / 445 Hoes Lane / P.O. Box 1331 / Piscataway, NJ 08855-1331, USA. Telephone: +Intl.908-562-3966.

\twocolumn
\maketitle

\begin{abstract}
Old master drawings were mostly created step by step in several layers using different materials. To art historians and restorers, examination of these layers brings various insights into the artistic work process and helps to answer questions about the object, its attribution and its authenticity.  However, these layers typically overlap and are oftentimes difficult to differentiate with the unaided eye. For example, a common layer combination is red chalk under ink.

In this work, we propose an image processing pipeline that operates on hyperspectral images to separate such layers.  Using this pipeline, we show that hyperspectral images enable better layer separation than RGB images, and that spectral focus stacking aids the layer separation. In particular, we propose to use two descriptors in hyperspectral historical document analysis, namely hyper-hue and extended multi-attribute profile (EMAP). Our comparative results with other features underline the efficacy of the three proposed improvements.
\end{abstract}


%
\IEEEpeerreviewmaketitle

\section{Introduction}

Red chalk was a highly popular drawing material until the late nineteenth century
\cite{brachert2001lexikon,eastaugh2007pigment}. In the artistic work process,
it has oftentimes been used for creating a first sketch, in order to later
overdraw it with ink. For art historians today, these sketches provide insights
into the creation process of the art work. In particular, differences between
the underlying sketch and the overdrawn picture can indicate changes in the
direction of the work.

In this work, we investigate the particular case where red chalk is overdrawn by ink. 
A widely used technique to visualize structures below a layer of ink is to
image via infrared reflectography (IRR) the object in the infrared range, at wavelengths above
2000\,nm. In this regime, ink becomes transparent. However, this approach is
not applicable to make red chalk. Red chalk consists primarily of natural red
clay containing iron oxide, and the reflectance of red chalk at wavelengths
above 2000\,nm is very similar to the image carrier (i.e., the paper or
parchment). As a consequence, this range of wavelengths can not be used to
visualize over-painted strata of red chalk
\cite{burmester1986neue,mairinger2003strahlenuntersuchung}.
The difficulties of displaying and distinguishing the drawn strata by
conventional IRR, or with remission-spectroscopy
poses a significant challenge to recover the underlying substrate layers.
This is also shown in the comparative sequence of images from the apocryphal
Rembrandt drawings in Munich (visible spectrum versus infrared imaging),
published by Burmester and Renger
\cite[pp.~19-31,~Figs.~7a-21b]{burmester1986neue}.

In this work, we propose to close this diagnostic gap to visualize red chalk
below ink by using hyperspectral imaging together with a pattern recognition
pipeline.  There are many works in the literature that used hyperspectral
imaging for document analysis and proved its superiority to RGB imaging
\cite{pelagotti2008multispectral, lettner2008contrast, diem2007multi}.  Our
contributions are three-fold: We propose two descriptors for using in
hyperspectral historical document analysis, namely hyper-hue and extended
multi-attribute profile (EMAP), and we address a common artifact in
hyperspectral imaging called focus shifting, and propose spectral focus
stacking as its solution. We evaluate the proposed approaches on drawings that
are created to exactly mimic the original work process.


\section{Hyperspectral Descriptors for Sketch Layer Separation}
\subsection{Extended Multi-Attribute Profile (EMAP)}
%
%
%
%
Attribute profiles are popular tools in remote
sensing~\cite{breen1996attribute,dalla2010morphological}. The idea is to
abstract morphological operators like opening or closing from specific shapes
of structuring elements. The building blocks of attribute profiles are
attribute filters that operate on connected components (CC) of lower or equal
gray level intensities. On each CC in the image, an attribute $A$ (e.g., the
area, standard deviation, or diameter of the CC) is computed and compared to a
threshold $\lambda$. If $A(CC_i) \ge \lambda$, it is preserved. Otherwise, the
$i$-th CC is merged with the closest
neighboring CC. Analogously to classical morphological operators, attribute
thickening (denoted as $\phi^A_{\lambda}(f)$) is the process of merging the CCs of image
$f$ to neighboring CC with higher gray level. Attribute thinning (denoted as
$\gamma^A_{\lambda}(f)$) is the process of merging the CCs of image $f$ to neighboring CC
with lower gray level.

The attribute thinning profile of an image $f$, denoted by $\Pi({\gamma^A_{\lambda}})(f)$, is generated by concatenating series of attribute thinning with an increasing criterion size $\lambda$:
\begin{equation}
\Pi({\gamma^A_{\lambda}})(f)=\{\Pi({\gamma^A_\lambda}):\Pi({\gamma^A_\lambda}) =\gamma^A_{\lambda}(f),\forall \lambda \in [0,...,n] \}
\end{equation}

Analogously, attribute thickening profile of an image $f$, denoted by $\Pi(\Phi^A_{\lambda})(f)$, is generated by concatenating series of attribute thickenings with an increasing criterion size $\lambda$:

\begin{equation}
\Pi(\Phi^A_{\lambda})(f)=\{\Pi(\Phi^T_{\lambda}):\Pi(\Phi^T_{\lambda}) =\Phi^T_{\lambda}(f),\forall \lambda \in [0,...,n] \}
\end{equation}

The attribute profile (AP) is generated by concatenating series of attribute thickening and thinning profiles with an increasing criterion size $\lambda$:

\begin{equation}
AP(f)=\{ {\overset{\forall \lambda \in [\lambda_1,...,\lambda_n]}{\Pi(\Phi^T_{\lambda})(f)}},f,{\overset{\forall \lambda \in [\lambda_1,...,\lambda_n]}\Pi(\Phi^T_{\lambda}(f)}\}
\end{equation}

In the case of $\lambda=0$, $\Pi(\gamma^T_{0})=\Pi(\Phi^T_0)=f$. Therefore, attribute profile vector's size will be $2n+1$, i.e., $n$ for attribute thinning, $n$ for attribute closing and one for the original image. 

By using more than one attribute and concatenating the generated APs, multi-attribute profiles (MAPs) are generated. Finally, stacking the computed MAPs over each spectral channel of a multi-/hyper-spectral image results in the extended multi-attribute profile (EMAP). 
EMAPs use both spatial and spectral signatures of a hyperspectral image and are
capable of modeling and describing an image based on different attributes, e.g.
area, standard deviation and moment of the CCs. In this work, we used the same
attributes and threshold values as the work by Ghamisi et
al.~\cite{ghamisi2014automatic}.

\subsection{Hyper-Hue}

In RGB color space, pixels are 3-dimensional. 
In this cube, $(0,0,0)^T$ corresponds to black color and $(1,1,1)^T$ represents the white color. The vector connecting these two points, the diagonal of the cube, is called achromatic axis. By projecting all points in the RGB cube on a plane which is perpendicular to the achromatic axis and includes the point $(0,0,0)^T$, the so-called chromatic plane with regular hexagon shaped borders is formed.

For $n$-dimensional hyperspectral images, the same concept can be extended. Each point $\textbf{x}_i$ is represented by $n$ values. 
{\color{black}{Therefore, an $n$-dimensional hyperspectral image is defined as
\begin{equation}
f:\mathcal{D}_f \subset \mathbb{Z}^2\to[0,1]^n\enspace.
\end{equation}
}}

The vector connecting the $n$-dimensional points, let $(0,\dots,0)^T$ denote
the black in $n$ dimensions, which we call HyperBlack, and let analogously
denote $(1,\dots,1)^T$ HyperWhite. Let furthermore $\textbf{a}$
denote the achromatic hyper-axis, which is the normal vector of the
hyper-chromatic plane $\textbf{P}$. In order to mathematically define
$\textbf{P}$, we derive its spanning unit vectors. In
an $n$-dimensional space, $\textbf{P}$ is spanned by $n-1$ pairwise
perpendicular $n$-dimensional unit vectors,
$\{\textbf{u}_1,\textbf{u}_2,\dots,\textbf{u}_n\}^T$. The vectors $\textbf{u}_i$ have
the properties that
(1) they start from the point HyperBlack, (2) they are pairwise
perpendicular, (3) they are unit vectors and therefore their norm is $1$, (4)
the direction of $\textbf{u}_1$ points towards
the chromatic hyper-space, (5) $\textbf{u}_i$ are orthogonal to $\textbf{a}$.

Suppose the first $n-m$ elements of $\textbf{u}_i$ are $0$ and the remaining
$m$ elements are non-zero. From these $m$ elements, denote the first one as $a$
and the remaining elements as $b$. As it is derived in
\cite{liu2017transformation}, we obtain a basis for $\mathbf{P}$ by setting
$a=\frac{m-1}{\sqrt{m(m-1)}}$ and $b=\frac{-1}{\sqrt{m(m-1)}}$. The projection
of a hyperspectral point $\textbf{x}_j$ onto $\mathbf{P}$ is then

\begin{equation}
\textbf{c}_j = (\textbf{x}_j\cdot\textbf{u}_1)\textbf{u}_1 + (\textbf{x}_j\cdot\textbf{u}_2)\textbf{u}_2 + \dots + (\textbf{x}_j\cdot\textbf{u}_n)\textbf{u}_n\enspace.
\end{equation}

Liu et al. \cite{liu2017transformation} defined hyper-hue $\textbf{h}$,
saturation $S$ and intensity $I$ of a hyperspectral point \textbf{x} via its
projection $\textbf{c}$ as

\begin{equation}
\textbf{h} = \frac{\textbf{c}}{\left\Vert\textbf{c}\right\Vert}\enspace,
\end{equation}

\begin{equation}\label{S}
S = \frac{\left\Vert\textbf{c}\right\Vert}{{c}_{\mathrm{max}}} = \mathrm{max}\{{x}_1,\dots,{x}_n\} - \mathrm{min}\{{x}_1,\dots,{x}_n\}\enspace,
\end{equation}

\begin{equation}\label{I}
I = \frac{1}{n}({x}_1+\dots+{x}_n)\enspace.
\end{equation}

In this way, an extension of HSI color space is defined for hyperspectral images.

\section{Processing Pipeline}

\subsection{Sensitivity Normalization}
\begin{figure}[tb]
	
	\begin{minipage}[b]{0.19\linewidth}
		\centering
		\centerline{\includegraphics[width=1\linewidth]{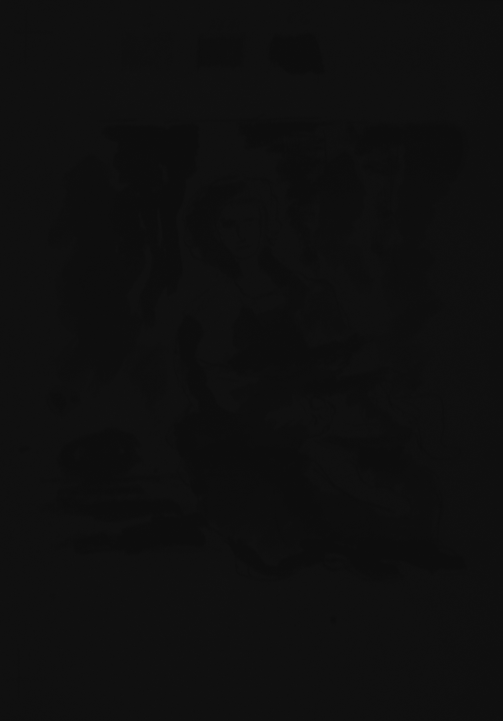}}
		\centerline{(a)}\medskip
	\end{minipage}
	\begin{minipage}[b]{0.19\linewidth}
		\centering
		\centerline{\includegraphics[width=1\linewidth]{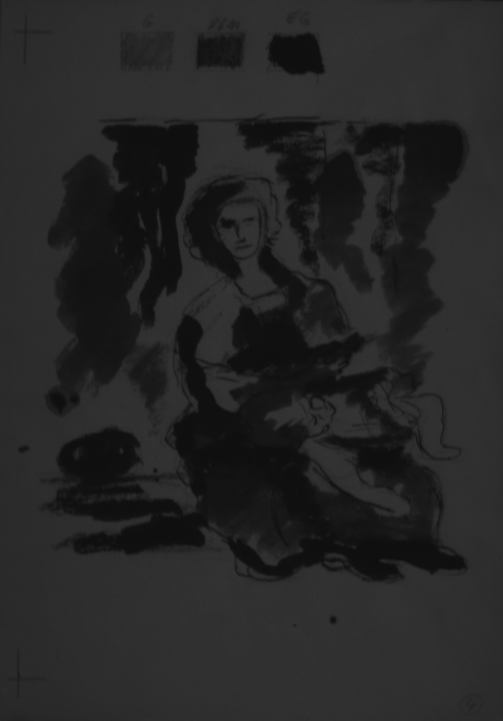}}
		\centerline{(b)}\medskip
	\end{minipage}
	\begin{minipage}[b]{0.19\linewidth}
		\centering
		\centerline{\includegraphics[width=1\linewidth]{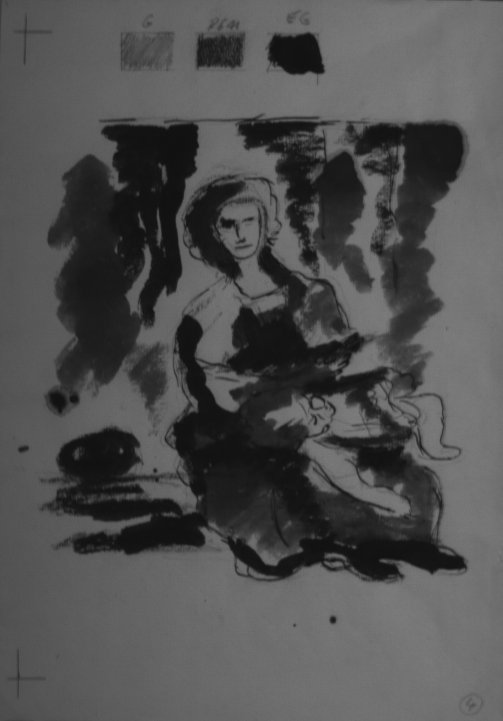}}
		\centerline{(c)}\medskip
	\end{minipage}
	\begin{minipage}[b]{0.19\linewidth}
		\centering
		\centerline{\includegraphics[width=1\linewidth]{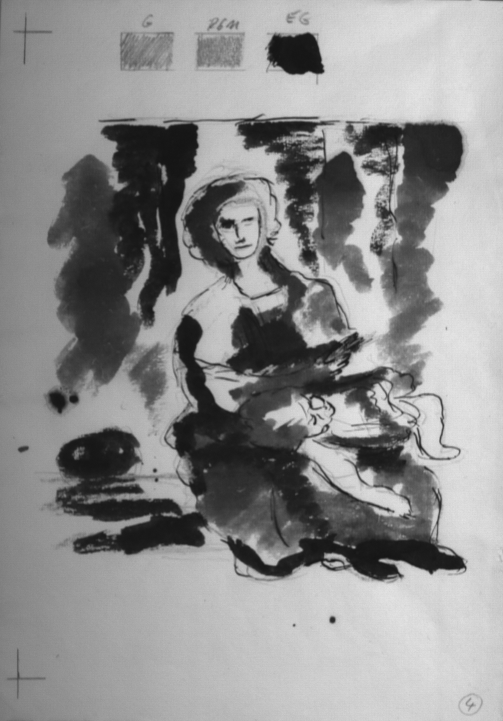}}
		\centerline{(d)}\medskip
	\end{minipage}
	\begin{minipage}[b]{0.19\linewidth}
		\centering
		\centerline{\includegraphics[width=1\linewidth]{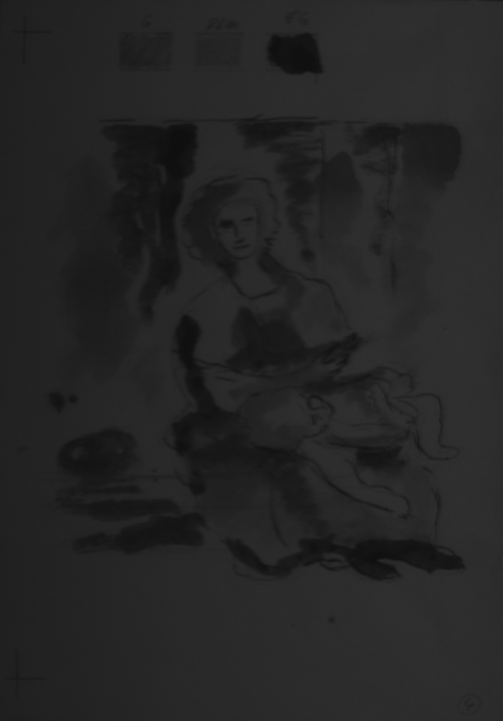}}
		\centerline{(e)}\medskip
	\end{minipage}
	\caption{Sample channels of the raw hyperspectral image. (a) channel 20, (b) channel 40, (c) channel 70, (d) channel 130, (e) channel 230. Comparing the channels, the uneven sensitivity of the camera sensor along the spectrum is obvious.}
	\label{Channels_OrigHSI}
\end{figure}

\begin{figure}[tb]
	
	\begin{minipage}[b]{0.32\linewidth}
		\centering
		\centerline{\includegraphics[width=1\linewidth]{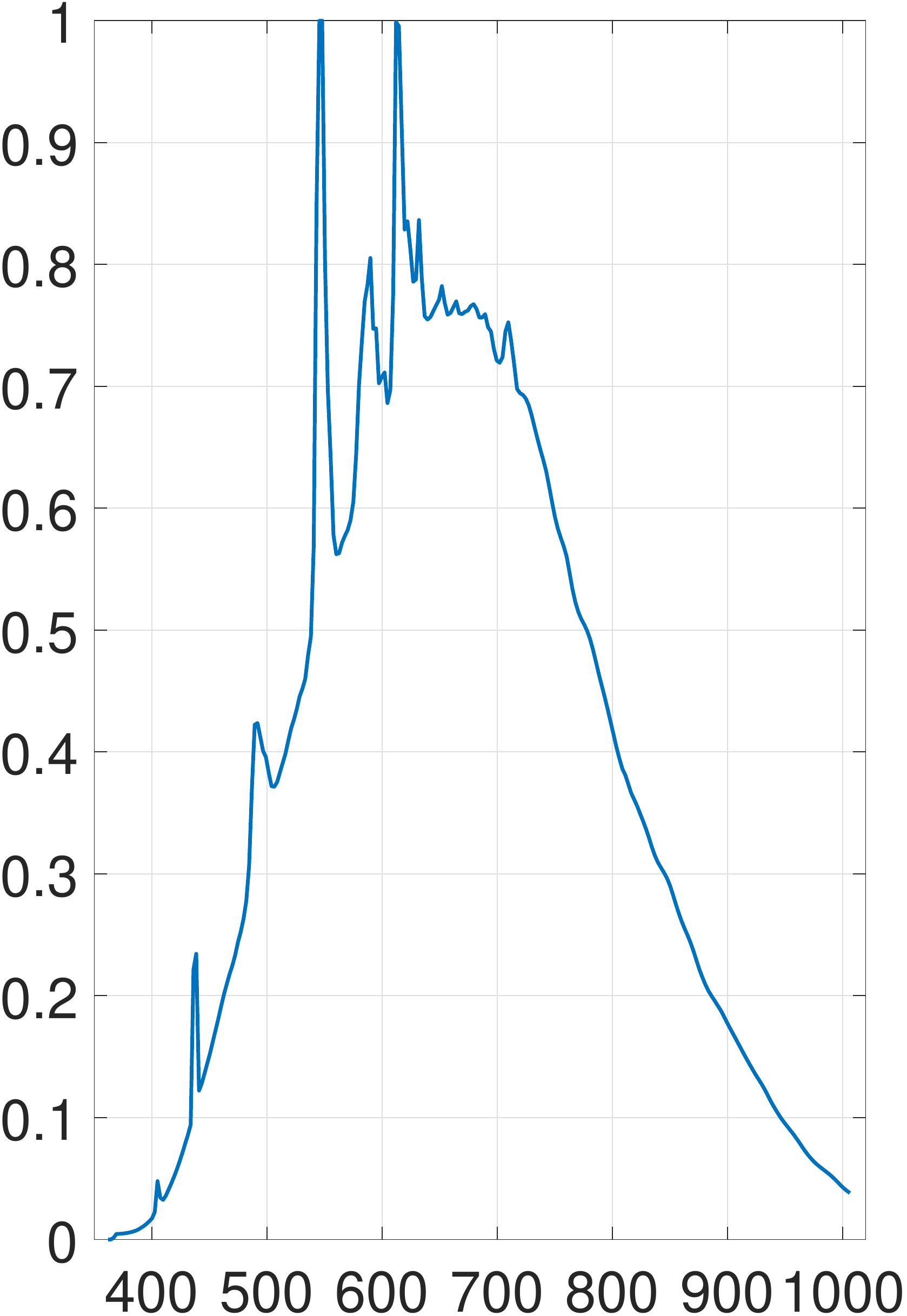}}
		\centerline{(a)}\medskip
	\end{minipage}
	\begin{minipage}[b]{0.32\linewidth}
		\centering
		\centerline{\includegraphics[width=1\linewidth]{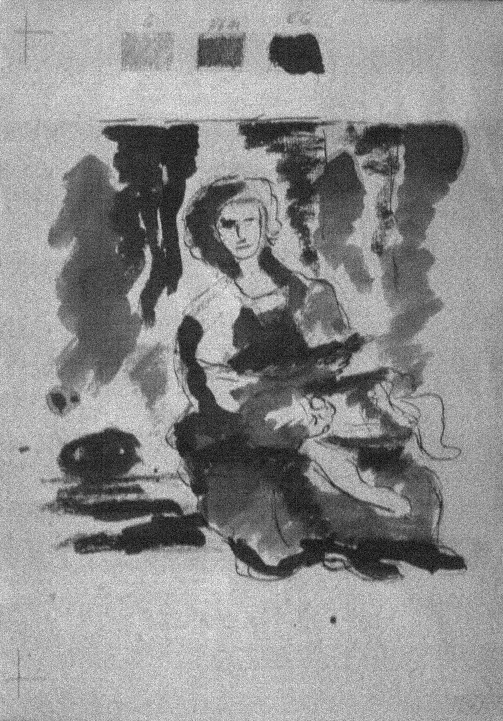}}
		\centerline{(b)}\medskip
	\end{minipage}
	\begin{minipage}[b]{0.32\linewidth}
		\centering
		\centerline{\includegraphics[width=1\linewidth]{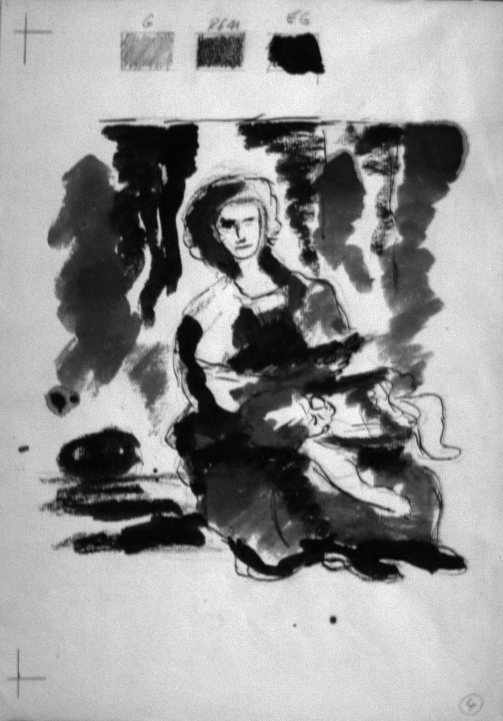}}
		\centerline{(c)}\medskip
	\end{minipage}
	\begin{minipage}[b]{0.32\linewidth}
		\centering
		\centerline{\includegraphics[width=1\linewidth]{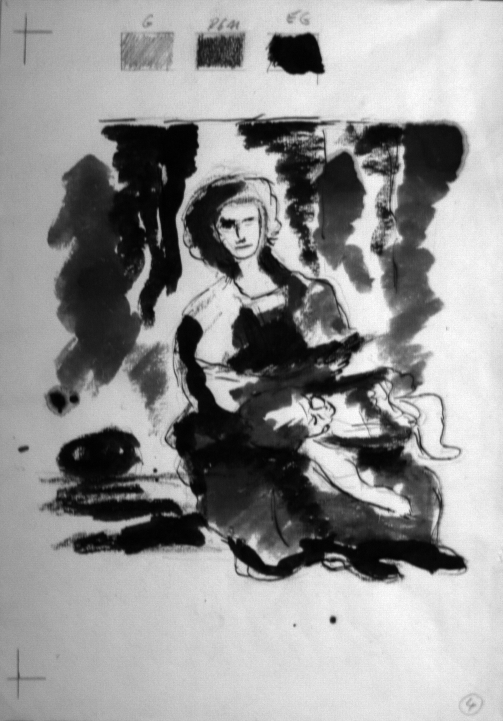}}
		\centerline{(d)}\medskip
	\end{minipage}
	\begin{minipage}[b]{0.32\linewidth}
		\centering
		\centerline{\includegraphics[width=1\linewidth]{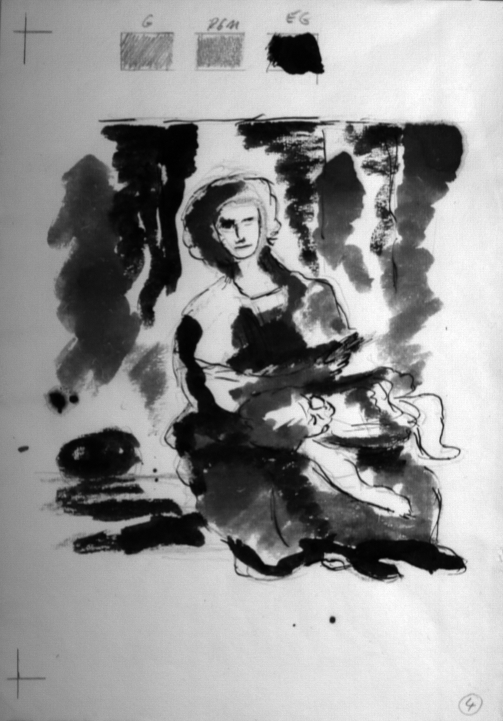}}
		\centerline{(e)}\medskip
	\end{minipage}
	\begin{minipage}[b]{0.32\linewidth}
		\centering
		\centerline{\includegraphics[width=1\linewidth]{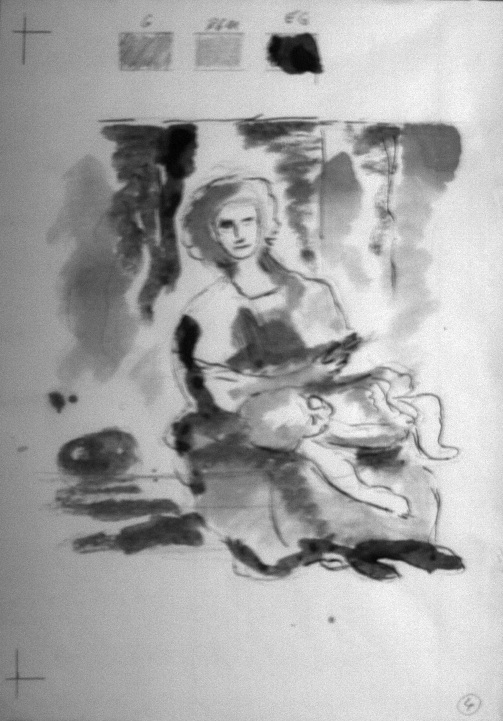}}
		\centerline{(f)}\medskip
	\end{minipage}
	\caption{Sensitivity-normalized sample channels of the raw hyperspectral image. (a) Normalized hyperspectral sensor sensitivity vs. wavelength (nm), (b) channel 20, (c) channel 40, (d) channel 70, (e) channel 130, (f) channel 230.}
	\label{Channels_OrigHSI_norm}
\end{figure}

Hyperspectral imaging setups suffer from various limitations and artifacts
which need to be corrected. 
Fig.~\ref{Channels_OrigHSI} (a)-(e) show five sample channels of a raw
hyperspectral image (HSI), namely channel 20 (representing 407.31\,nm
wavelength), channel 40 (representing 455.41\,nm wavelength), channel 70
(representing 528.32\,nm wavelength), channel 130 (representing 676.93\,nm
wavelength), channel 230 (representing 932.82\,nm wavelength). As it can be
observed, the sensitivity of the HS camera sensor along the spectrum is not
uniform. Using a white reference, the uneven sensitivity can be corrected.
Fig.~\ref{Channels_OrigHSI_norm}-(a) shows the normalized sensitivity diagram
of the sensor measured from a white reference. The inverse of this diagram is used
as the sensitivity normalization coefficient.
Fig.~\ref{Channels_OrigHSI_norm} (b)-(f) show the sensitivity-normalized
version of the channels presented in Fig.~\ref{Channels_OrigHSI} (a)-(e),
respectively.

\begin{figure}[tb]
	
	\begin{minipage}[b]{0.32\linewidth}
		\centering
		\centerline{\includegraphics[width=1\linewidth]{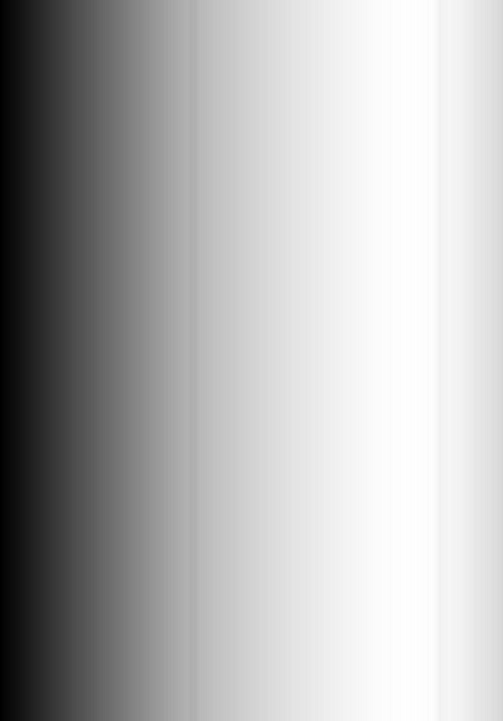}}
		\centerline{(a)}\medskip
	\end{minipage}
	\begin{minipage}[b]{0.32\linewidth}
		\centering
		\centerline{\includegraphics[width=1\linewidth]{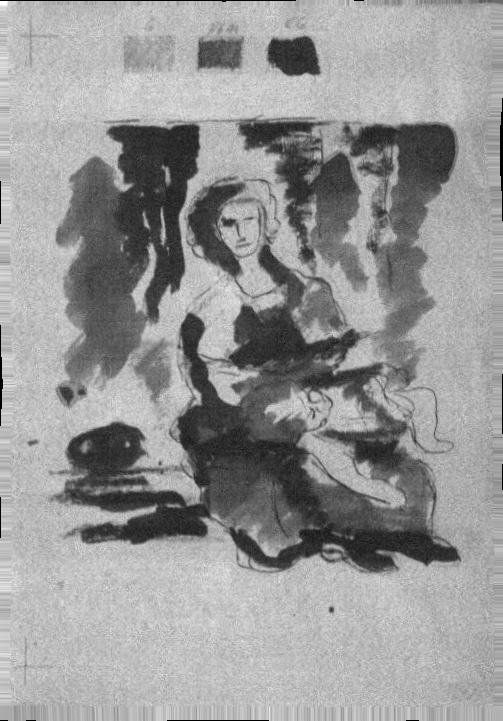}}
		\centerline{(b)}\medskip
	\end{minipage}
	\begin{minipage}[b]{0.32\linewidth}
		\centering
		\centerline{\includegraphics[width=1\linewidth]{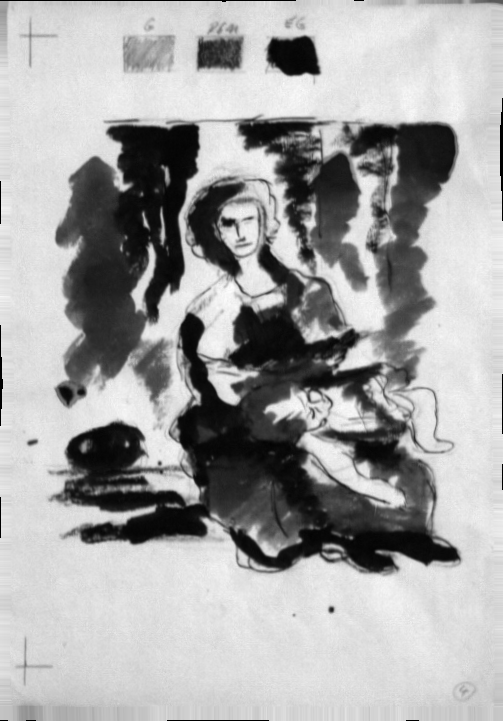}}
		\centerline{(c)}\medskip
	\end{minipage}
	\begin{minipage}[b]{0.32\linewidth}
		\centering
		\centerline{\includegraphics[width=1\linewidth]{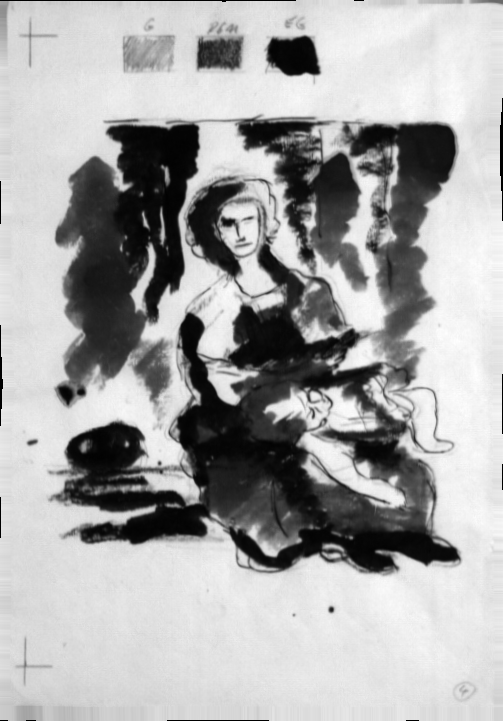}}
		\centerline{(d)}\medskip
	\end{minipage}
	\begin{minipage}[b]{0.32\linewidth}
		\centering
		\centerline{\includegraphics[width=1\linewidth]{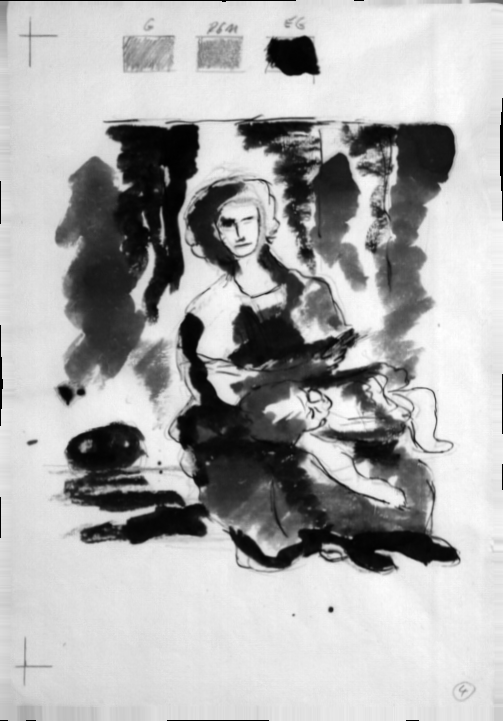}}
		\centerline{(e)}\medskip
	\end{minipage}
	\begin{minipage}[b]{0.32\linewidth}
		\centering
		\centerline{\includegraphics[width=1\linewidth]{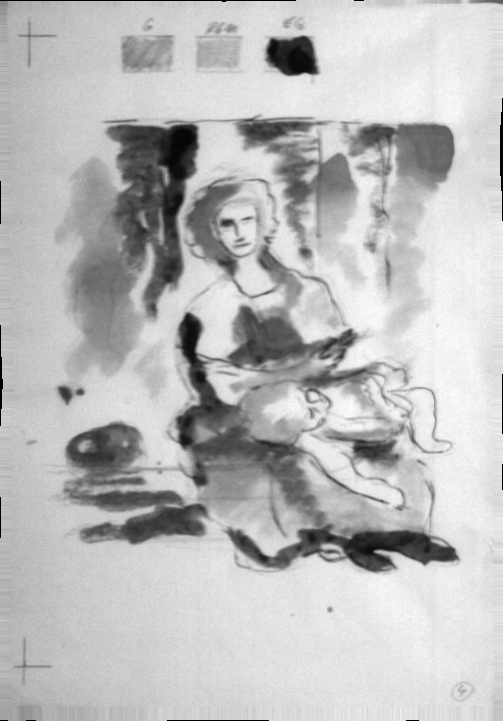}}
		\centerline{(f)}\medskip
	\end{minipage}
	\caption{Illumination-corrected and registered sample channels of the raw hyperspectral image to the ground truth. (a) Normalized uneven illumination field, (b) channel 20, (c) channel 40, (d) channel 70, (e) channel 130, (f) channel 230.}
	\label{Channels_OrigHSI_NormRegIllum}
\end{figure}

Every imaging setup needs good lighting for an acceptable acquisition and HS imaging is not an exception. In real world scenario, in a museum for instance, the subject may not be evenly illuminated. To simulate this situation, we sidelit our scene. Using a white reference, we estimate the uneven lighting, as shown in Fig.~\ref{Channels_OrigHSI_NormRegIllum}-(a). Fig.~\ref{Channels_OrigHSI_NormRegIllum} (b)-(f) show the illumination-corrected version of the Fig.~\ref{Channels_OrigHSI_norm} (b)-(f), respectively.

\subsection{Focus Stacking} 
\begin{figure}[tb]
	
	\begin{minipage}[b]{0.24\linewidth}
		\centering
		\centerline{\includegraphics[width=1\linewidth]{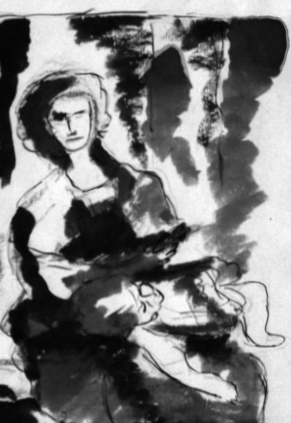}}
		\centerline{(a)}\medskip
	\end{minipage}
	\begin{minipage}[b]{0.24\linewidth}
		\centering
		\centerline{\includegraphics[width=1\linewidth]{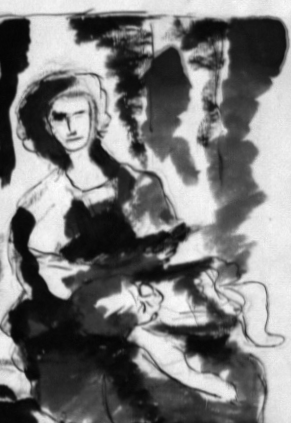}}
		\centerline{(b)}\medskip
	\end{minipage}
	\begin{minipage}[b]{0.24\linewidth}
		\centering
		\centerline{\includegraphics[width=1\linewidth]{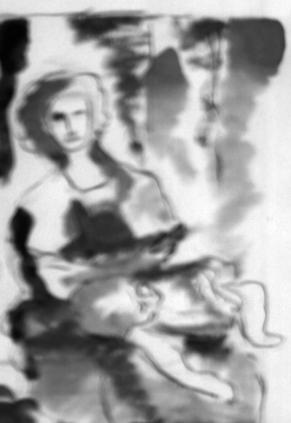}}
		\centerline{(c)}\medskip
	\end{minipage}
	\begin{minipage}[b]{0.24\linewidth}
		\centering
		\centerline{\includegraphics[width=1\linewidth]{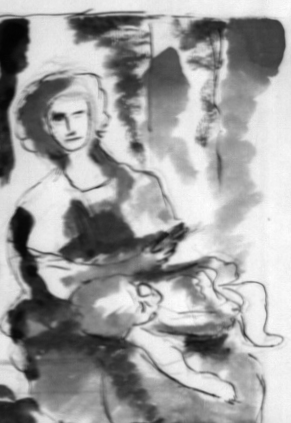}}
		\centerline{(d)}\medskip
	\end{minipage}
	\caption{$H_1$ which is focused on the blue spectrum vs. $H_2$ which is focused on the red spectrum. (a) $H_1$ channel 41 (457.82\,nm), (b) $H_2$ channel 41 (457.82\,nm), (c) $H_1$ channel 200 (854.97\,nm), (d) $H_2$ channel 200 (854.97\,nm).}
	\label{FocusStacking_Channels}
\end{figure}

Common HS cameras suffer from focus shifting, which is a well-known artifact in
the field. It leads to the issue that not all of the channels are
simultaneously in focus when making a multispectral acquisition.
Fig.~\ref{FocusStacking_Channels} shows this behavior for two hyperspectral
images, namely $H_1$ and $H_2$. For acquiring $H_1$, the lens is focused with
the blue range aimed to be in focus. $H_2$, on the other hand, is captured by
having the red spectrum in focus. Fig.~\ref{FocusStacking_Channels} (a) shows
channel 41, representing 458.82\,nm wavelength, of $H_1$.
Fig.~\ref{FocusStacking_Channels} (b) shows the same channel of $H_2$.
Especially on fine edges, we can observe that (a) is sharper and more in focus.
Similarly, Fig.~\ref{FocusStacking_Channels} (c) and (d) show the channel 200,
representing the 854.97\,nm wavelength, of $H_1$ and $H_2$, respectively. This
time the channel corresponding to $H_2$ is sharper than $H_1$.

One contribution of this work lies in producing one hyperspectral image with
all channels in focus via spectral focus stacking. To this end, we acquire
two images with two different focus points, one in the blue spectrum and one in
the red spectrum. The final all-in-focus image is generated from the in-focus
channels of the two input images. In our work, we generate our final
all-in-focus image by using the first 75 channels from $H_1$ and the remaining
183 channels from $H_2$. We quantitatively compared our all-in-focus HSI with
$H_1$ and $H_2$. The results are presented in Table~\ref{HSI_all_in_focus} and
are discussed in Sec.\ref{sec:results}.


\subsection{Classification}
In a previous study on this application we followed an
unsupervised approaches with k-means and GMM clustering algorithms~\cite{Davari17a}, which
performed weakly, especially for diluted red chalk. In this work, we assume
that it is feasible to obtain a limited number of labeled pixels by a
specialist, e.g., an art historian.  This allows to use supervised learning to
evaluate the proposed features and processing pipeline for layer separation.
We consider the three classes red chalk, diluted red chalk and black ink.
Classification is performed using a random forest (RF), with 10 trees. The
number of variables for training the trees and bagging is set to the square
root of the number of features, as proposed by
Breiman~\cite{breiman2001random}. We used 100 random samples per class for
training and the rest for testing. We repeated this process 25 times and
reported the average classification performance metrics and their standard
deviation (SD).
In our dataset, the number of pixels for these classes is $10791$, $23528$ and
$85000$, respectively. For training, we select $100$ pixels from each class,
which corresponds to $0.9\%$, $0.4\%$ and $0.1\%$ of each class, respectively.

\section{Evaluation}\label{sec:evaluation}

\subsection{Dataset}
\subsubsection{Phantom Data}
\begin{figure}[tb]
	
	\begin{minipage}[b]{0.32\linewidth}
		\centering
		\centerline{\includegraphics[width=1\linewidth]{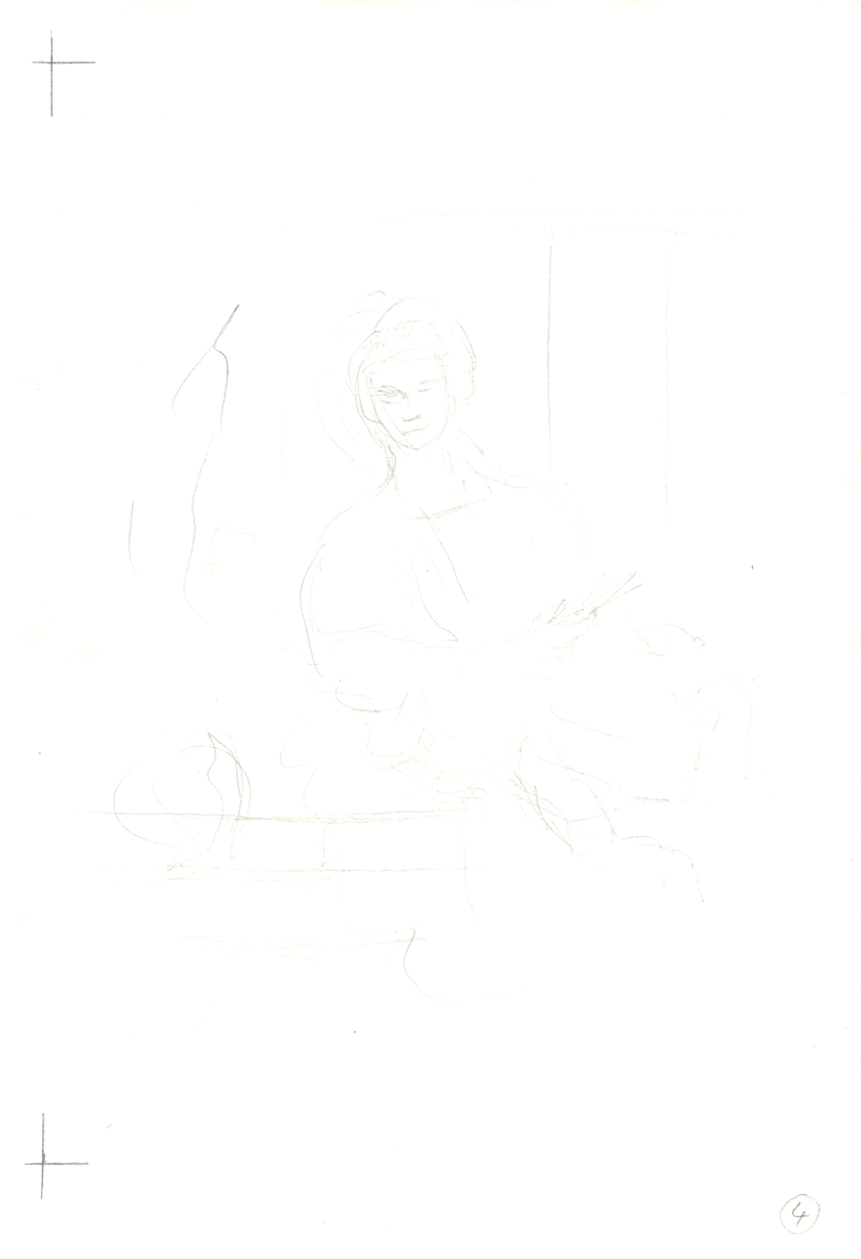}}
		\centerline{(a)}\medskip
	\end{minipage}
	\begin{minipage}[b]{0.32\linewidth}
		\centering
		\centerline{\includegraphics[width=1\linewidth]{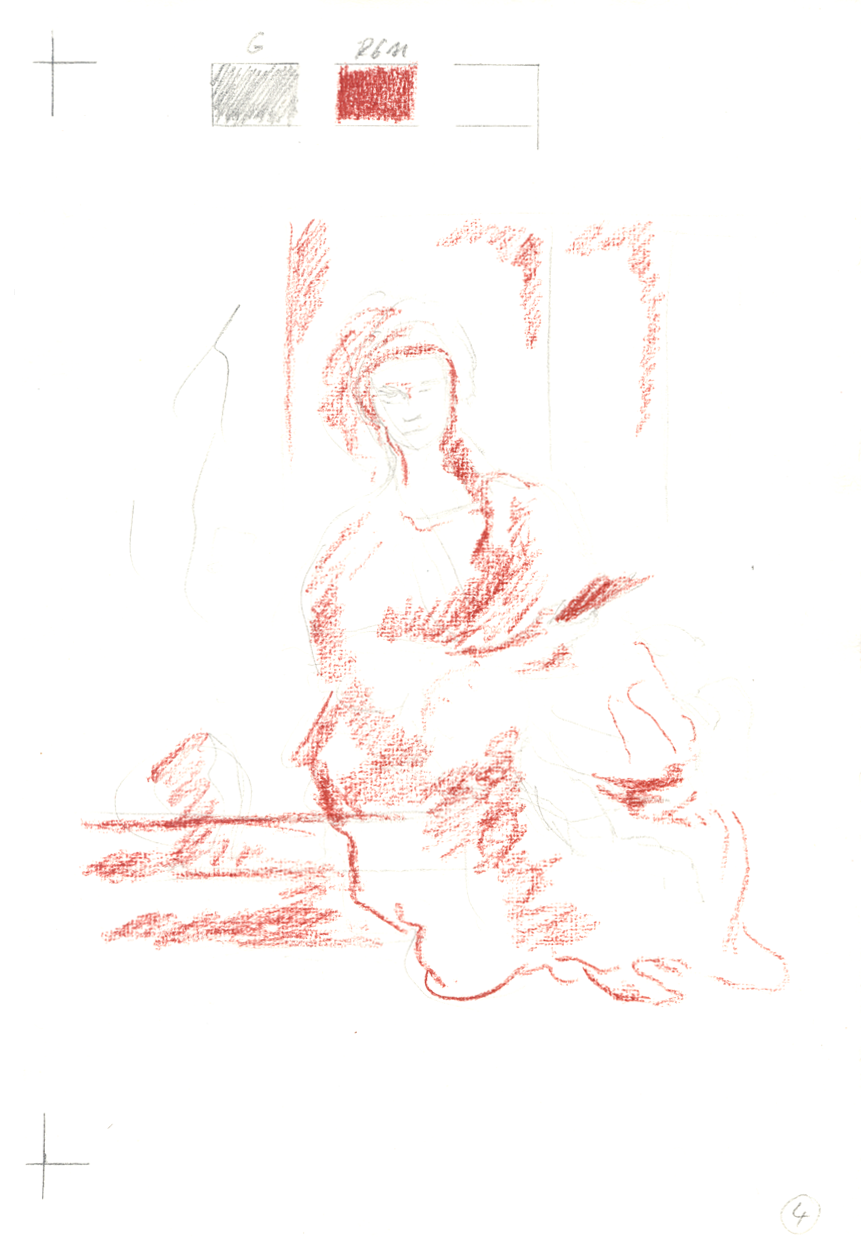}}
		\centerline{(b)}\medskip
	\end{minipage}
	\begin{minipage}[b]{0.32\linewidth}
		\centering
		\centerline{\includegraphics[width=1\linewidth]{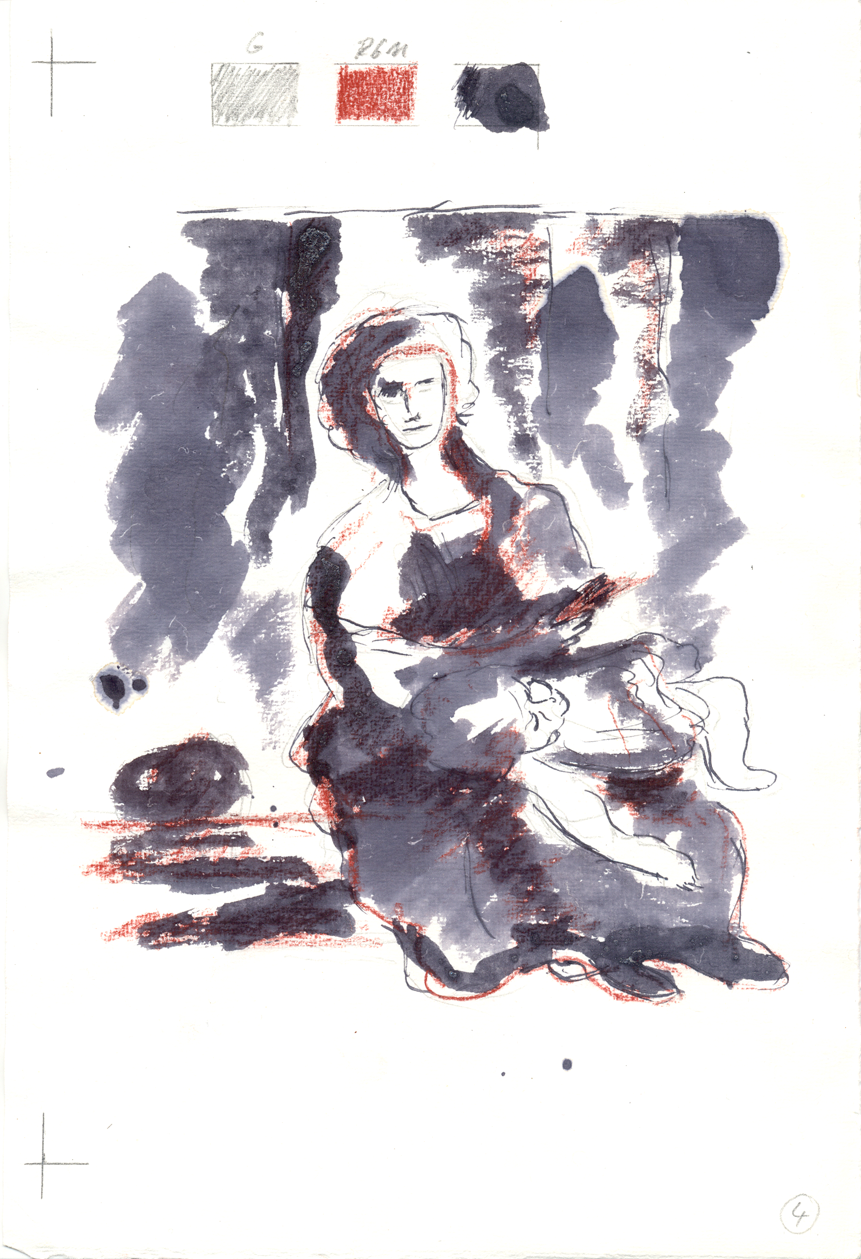}}
		\centerline{(c)}\medskip
	\end{minipage}

	\caption{Sample layers from the data of the creation process as basis of evaluation: (a) Step 1: first graphite sketch. (b) Step 2: underdrawings with red chalk. (c) Steps 3 and 4: drawing with pen and iron gall ink plus final wash with two dilutions of ink in ``two bowl technique'', as described by Armenini \cite{armenini1586veri}[pp.54-55]. and Meder \cite{meder1919handzeichnung}[pp.54-55]. Delineation after: Stefano della Bella, ``Mother with two children'', Florence, Galleria degli Uffici, Gabinetto Disegni e stampe, Inv.-Nr. 5937S.}
	\label{DataLayers}
\end{figure}

We created a set of sketches with multiple layers of graphite, chalk, and
different inks of the same chemical composition that were commonly used in old
master drawings. After each layer was drawn, the picture was scanned with a
book scanner (Zeutschel OS 12000, in RGB mode). This step-by-step documentation
of the controlled creation process allows to compute ground truth drawing
layers, by subtracting two subsequent scanned images. A sample sketch from this
data is shown in Figure \ref{DataLayers}.

\subsubsection{Hyperspectral Imaging}
\begin{figure}[tb]
	\centering
	\centerline{\includegraphics[width=.8\linewidth]{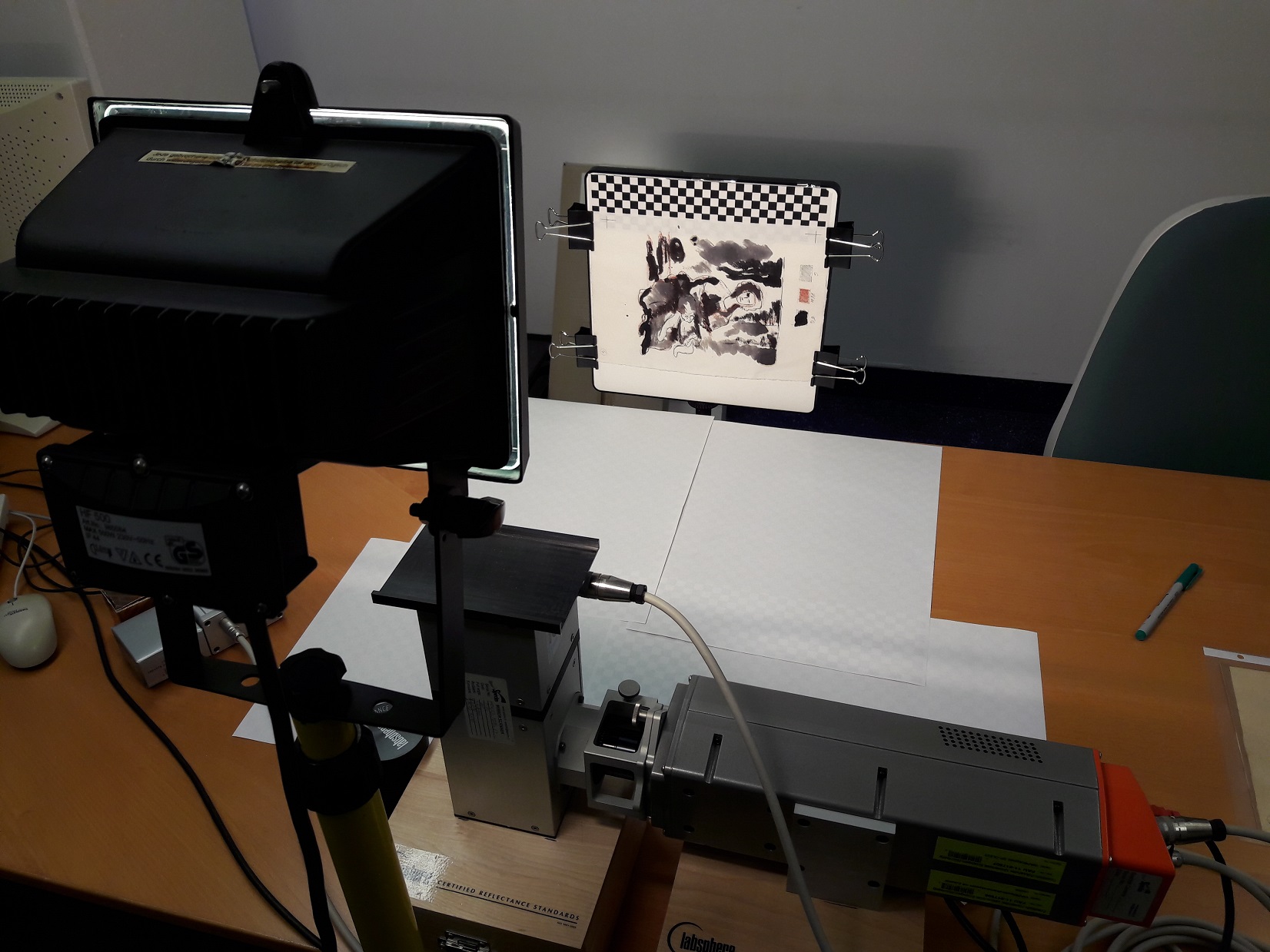}}
	\caption{Imaging Setup.}
	\label{Imaging_Setup}
\end{figure}

For imaging, we use a Specim PFD-CL-65-V10E hyperspectral camera equipped with
a CMOS sensor, capable of capturing the spectrum in a wavelength range of
400\,nm to 1000\,nm. The imaging setup is shown in Fig.~\ref{Imaging_Setup}.
We use a lens with 16\,mm focal length and the distance between the subject and
the camera is 68\,cm. The document is illuminated with a 500\,W tungsten lamp.

\subsubsection{Simulated RGB}
\begin{figure}[tb]
	
	\begin{minipage}[b]{0.32\linewidth}
		\centering
		\centerline{\includegraphics[width=1\linewidth]{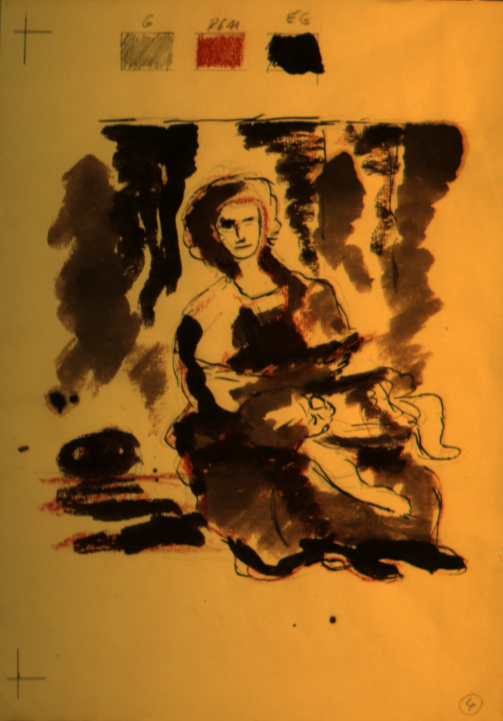}}
		\centerline{(a)}\medskip
	\end{minipage}
	\begin{minipage}[b]{0.32\linewidth}
		\centering
		\centerline{\includegraphics[width=1\linewidth]{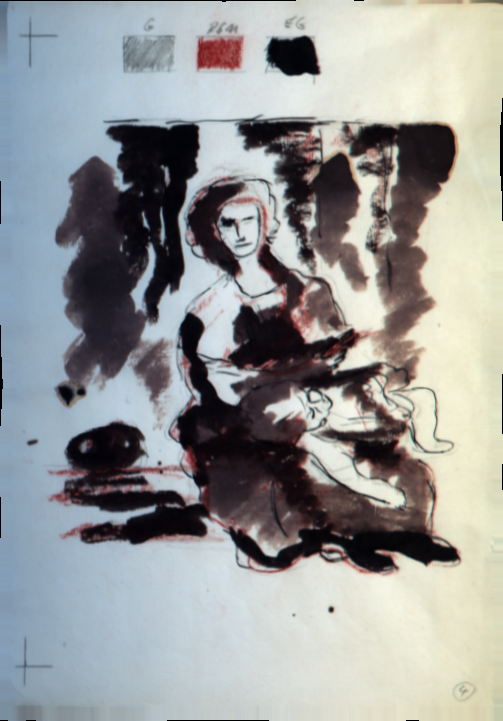}}
		\centerline{(b)}\medskip
	\end{minipage}
	\begin{minipage}[b]{0.32\linewidth}
		\centering
		\centerline{\includegraphics[width=1\linewidth]{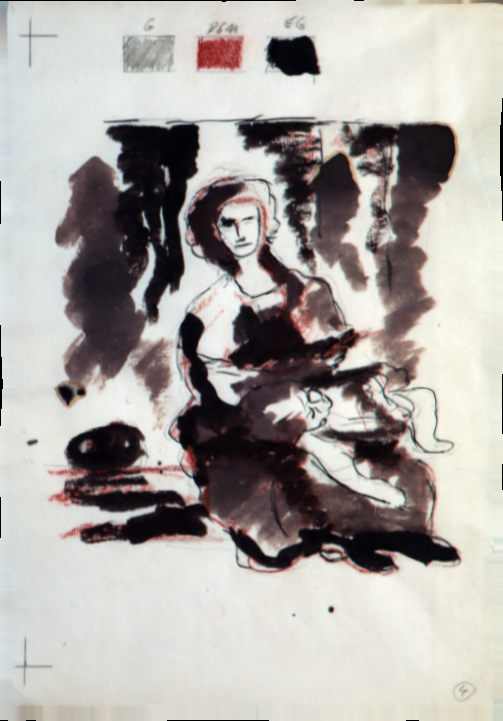}}
		\centerline{(c)}\medskip
	\end{minipage}
	\caption{Simulated RGB image from the hyperspectral image. (a) RGB image generated from the raw HSI in Fig.~\ref{Channels_OrigHSI}, (b) RGB image generated from the sensitivity-normalized HSI in Fig.~\ref{Channels_OrigHSI_norm}, (c) RGB image generated from illumination-corrected HSI in Fig.~\ref{Channels_OrigHSI_NormRegIllum}. }
	\label{Fake_RGB}
\end{figure}

In order to compare the effectiveness of using hyperspectral images for
layer separation with RGB images, we simulated RGB images from our
hyperspectral images. The blue color in RGB domain is corresponding to the
wavelengths between 415\,nm and 495\,nm (HSI channels 24 to 56). Similarly, the
green color corresponds to the wavelengths range of 495\,nm-570\,nm (HSI channels
57 to 87) and the red color lies between 620\,nm-750\,nm (HSI channels 108 to 156).
We generated the red, green and blue channels by taking the average of HSI
channels 108-156, 57-87 and 24-56, respectively. The reason that we do not use
the RGB image acquired by the board scanner for comparison is that the board
scanner has newer sensor, higher resolution, higher signal to noise ratio (SNR)
and better lighting condition. Therefore, the comparison would not be fair.
Fig.\ref{Fake_RGB} shows the simulated RGBs from the HSIs, before and after
pre-processing. 

\subsection{Evaluation Protocol}
\subsubsection{Registration of HSI to the ground truth}
Our ground truth, generated from Fig.~\ref{DataLayers}, is acquired by a board scanner. The HSI images are acquired via a line scanner hyperspectral camera. Different modalities, resolutions, aspect ratios and the non-flat surface of the paper make the images from these modalities geometrically different. In order to compare the HSI analysis output, hyperspectral images need to be registered to the board scanner image. In a previous study~\cite{Davari17b}, we concluded that a non-rigid registration using residual complexity similarity measure (RC) \cite{myronenko2010intensity} suits our purpose well. Therefore, we use RC to register our HSI to the RGB image acquired by the board scanner.

\subsubsection{Metrics}
To evaluate the classification performances, we used overall accuracy (OA), average accuracy (AA) and Kappa coefficient metrics. OA is the number of correctly classified instances divided by the number of all samples, while AA is the mean class-based accuracies. 
The kappa statistic is a measure of how closely the classified samples matches the ground truth. 
By measuring the expected accuracy, it results in a statistic expressing the accuracy of a random classifier.

\subsection{Results}\label{sec:results}

\subsubsection{Impact of Spectral Focus Stacking}
In order to study the effect of spectral focus stacking, we conducted two sets
of experiments on simulated RGB images and HSI images. In the first experiment,
we generated RGB images from $H_1$, $H_2$ and all-in-focus HSI. In the second
experiment, we carried out the classification on the illumination-corrected
$H_1$, $H_2$ and all-in-focus HSI. The results for these two experiments are
presented in Table~\ref{HSI_all_in_focus}. It can be seen that in both
scenarios, spectral focus stacking yields better AA, OA and Kappa performance. 

\begin{table}[t]
	\centering
	\caption{Spectral focus stacking results.}
	\begin{tabular}{|@{ }l@{ }|@{ }ccc@{ }|}
		\hline
		Feature & AA\% ($\pm$SD) & OA\% ($\pm$SD) & Kappa ($\pm$SD) \\
		\hline
		\rowcolor[rgb]{ .906,  .902,  .902} \multicolumn{4}{|c|}{Simulated RGB image from HSI} \\
		\hline
		$H_1$  & 70.63 ($\pm$1.41)  & 60.82 ($\pm$2.51)  & 0.3515 ($\pm$0.0227) \\
		$H_2$  & 72.32 ($\pm$1.15)  & 63.62 ($\pm$3.09)  & 0.3777 ($\pm$0.0313) \\
		Focus Stacking & \textbf{73.72} ($\pm$1.10)  & \textbf{64.96} ($\pm$2.40)  & \textbf{0.3980} ($\pm$0.0257) \\
		\hline
		\rowcolor[rgb]{ .906,  .902,  .902} \multicolumn{4}{|c|}{Illumination-corrected HSI} \\
		\hline
		$H_1$  & 74.76 ($\pm$0.94)  & 64.67 ($\pm$1.45)  & 0.3998 ($\pm$0.0169) \\
		$H_2$  & 76.12 ($\pm$0.96)  & 66.34 ($\pm$1.42)  & 0.4186 ($\pm$0.0165) \\
		Focus Stacking & \textbf{76.57} ($\pm$0.94)  & \textbf{67.21} ($\pm$3.56)  & \textbf{0.4304} ($\pm$0.0366) \\
		\hline
	\end{tabular}%
	\label{HSI_all_in_focus}%
\end{table}%

\subsubsection{Layer Separation Performance of the Proposed Features}
As spectral focus stacking results in better performance, the remaining
computations are performed over all-in-focus images.
To study the impact of illumination correction, hyper-hue, and EMAP, we
generated the following features. 
\begin{enumerate}
	\item \label{1} SimRGB: Simulated RGB image, generated from the illumination-uncorrected all-in-focus HSI, 
	\item \label{2} SimRGB-IC: Simulated RGB image, generated from the illumination-corrected HSI, 
	\item \label{2_SI} SimRGB-IC-SI: SimRGB-IC, saturation ($S$ in Eq.~\ref{S}) and illumination ($I$ in Eq.~\ref{I}) concatenated together,
	\item \label{2_EMAP} SimRGB-IC-EMAP: EMAP computed on SimRGB-IC. We used area as the only EMAP attribute with 20 thresholds $\lambda$. In order to choose the threshold values, we followed a similar approach to Ghamisi et al. \cite{ghamisi2014automatic},
	\item \label{3_0} HSI: Illumination-uncorrected all-in-focus HSI,
	\item \label{3} HSI-IC: Illumination-corrected all-in-focus HSI,
	\item \label{4} HSI-DR: HSI-IC projected to its PCA components such that 99.9\% of its variance are preserved,
	\item \label{5} HSI-h: Hyper-hue computed from the illumination-corrected HSI,
	\item \label{6} HSIhSI: HSI-IC, hyper-hue, saturation ($S$) and illumination ($I$) concatenated together,
	\item \label{7} HSIhSI-DR: Dimensionality reduced HSIhSI via PCA so that 99.9\% of its variance is preserved.
	\item \label{8} HSI-EMAP: EMAP computed on dimensionality reduced HSI-IC: EMAP's parameters are chosen similar to SimRGB-IC-EMAP,
	\item \label{9} HSIhSI-EMAP: EMAP computed on dimensionality reduced HSIhSI. EMAP parameters are chosen similar to SimRGB-IC-EMAP.
\end{enumerate}

The results for the features are presented in
Table~\ref{Overall_Res}. The first observation is that illumination correction
always improves the results, both for SimRGB vs. SimRGB-IC and for HSI vs.
HSI-IC. Furthermore, comparing the SimRGB-IC with HSI-IC indicates that an
illumination-corrected HSI performs better than an RGB image. In HSI-DR,
applying PCA further improves the HSI performance.

Hyper-hue computed over HSI (HSI-h) results in a big jump in performance.
Furthermore, the standard deviation in HSI-h is smallest among all the other
features which indicates a high stability of this feature.
Combining hyper-hue, saturation and illumination on the HSI image (HSI-hSI) can
not exceed the performance of hyper-hue alone. Also dimensionality reduction on
this combination (HSI-hSI-DR) can not compete with HSI-h, and performs even
worse than HSI-hSI.
EMAP computed on the HSI (HSI-EMAP) results in a performance that is well
comparable with HSI-h. Finally, computing EMAP over HSIhSI leads to a slight
improvement and results in the overall best layer separation performance.

It is worth mentioning that the threshold values we choose for EMAP by
following the proposed method in \cite{ghamisi2014automatic} are probably not optimal.
Observing a competitive performance by EMAP in this work motivates us to study
other attributes and threshold values in our future works.

\begin{table}[t]
	\centering
	\caption{Performances of the features \ref{1} to \ref{9}.}
	\begin{tabular}{|@{ }l@{ }|@{ }c@{ }c@{ }c@{ }|}
		\hline
		Feature   & AA\% ($\pm$SD)    & OA\% ($\pm$SD)    & Kappa ($\pm$SD) \\
		\hline
		SimRGB     & 71.83 ($\pm$0.79)  & 62.05 ($\pm$1.90)  & 0.3632 ($\pm$0.0178) \\
		SimRGB-IC  & 73.72 ($\pm$1.10)  & 64.96 ($\pm$2.40)  & 0.3980 ($\pm$0.0257) \\
		SimRGB-IC-SI & 74.29 ($\pm$0.61) &  66.08 ($\pm$2.57)  &  0.4119 ($\pm$0.0261) \\
		SimRGB-IC-EMAP & 74.63 ($\pm$0.77) &  67.25 ($\pm$1.84)  &  0.4251 ($\pm$0.0170) \\
		\hline
		HSI & 75.43 ($\pm$1.05) & 66.94 ($\pm$2.11) &   0.4196 ($\pm$0.0217) \\
		HSI-IC     & 76.57 ($\pm$0.94)  & 67.21 ($\pm$3.56)  & 0.4304 ($\pm$0.0366) \\
		\hline
		HSI-DR   & 80.35 ($\pm$0.66)  & 72.58 ($\pm$1.53)  & 0.5019 ($\pm$0.0183) \\
		HSI-h      & \textbf{83.00} ($\pm$0.47)  & \textbf{77.39} ($\pm$1.28)  & \textbf{0.5731} ($\pm$0.0161) \\
		HSIhSI     & 82.86 ($\pm$0.52)  & 77.16 ($\pm$1.53)  & 0.5701 ($\pm$0.0213) \\
		HSIhSI-DR  & 79.58 ($\pm$0.86)  & 71.00 ($\pm$2.41)  & 0.4817 ($\pm$0.0273) \\
		HSI-EMAP   & \textbf{82.61} ($\pm$1.11)  & \textbf{77.35} ($\pm$2.53)  & \textbf{0.5719} ($\pm$0.0350) \\
		HSIhSI-EMAP & \textbf{83.08} ($\pm$0.89)  & \textbf{77.70} ($\pm$1.18)  & \textbf{0.5766} ($\pm$0.0191) \\
		\hline
	\end{tabular}%
	\label{Overall_Res}%
\end{table}%

In order to see the effect of a better sensor, SNR and lighting, we also
classified on the down-sampled RGB image that is acquired by the RGB board scanner (see 
Table~\ref{RGBResult_BoardScanner}). These results
are superior to the HSI-based
results, which was expected. Along with the results in
Table~\ref{Overall_Res}, we conclude that 
multi-/hyper-spectral imaging with suitable processing can outperform RGB
imaging when operating on images with identical noise and photon statistics.
While the photon statistics is typically bounded by the fact that historic
documents may not be exposed to too much light, it will be interesting to
investigate multi-spectral imaging with a DSLR camera due to the improved
resolution, SNR, and dynamic range in future work.

\begin{table}[t]
	\centering
	\caption{Down-sampled board scanner-acquired RGB image performance.}
	\begin{tabular}{|@{ }ccc@{ }|}
		\hline
		AA\% ($\pm$SD)    & OA\% ($\pm$SD)    & Kappa ($\pm$SD) \\
		\hline
		87.71 ($\pm$0.66)  & 83.68 ($\pm$1.31)  & 0.6773 ($\pm$0.0206) \\
		\hline
	\end{tabular}%
	\label{RGBResult_BoardScanner}%
\end{table}%

\begin{figure}[t]
	
	\begin{minipage}[b]{0.325\linewidth}
		\centering
		\centerline{\includegraphics[width=1\linewidth]{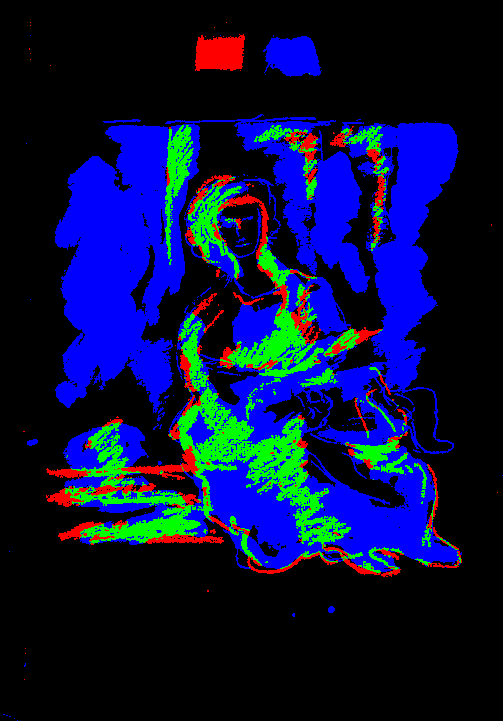}}
		\centerline{(a)}\medskip
	\end{minipage}
	%
	\begin{minipage}[b]{0.325\linewidth}
		\centering
		\centerline{\includegraphics[width=1\linewidth]{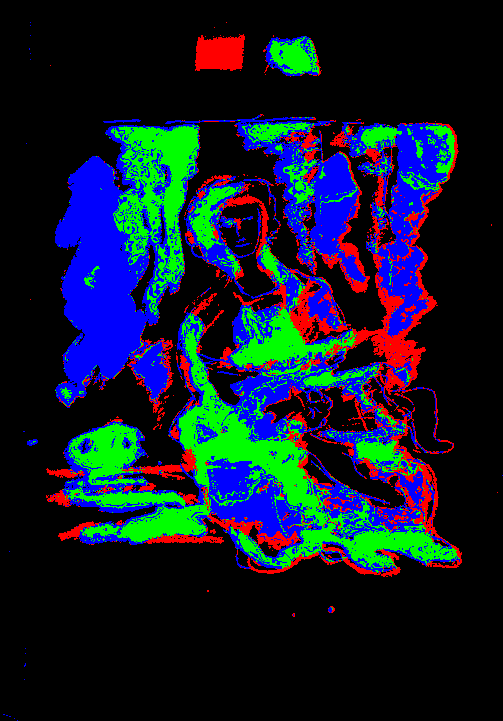}}
		\centerline{(b)}\medskip
	\end{minipage}
	%
	\begin{minipage}[b]{0.325\linewidth}
		\centering
		\centerline{\includegraphics[width=1\linewidth]{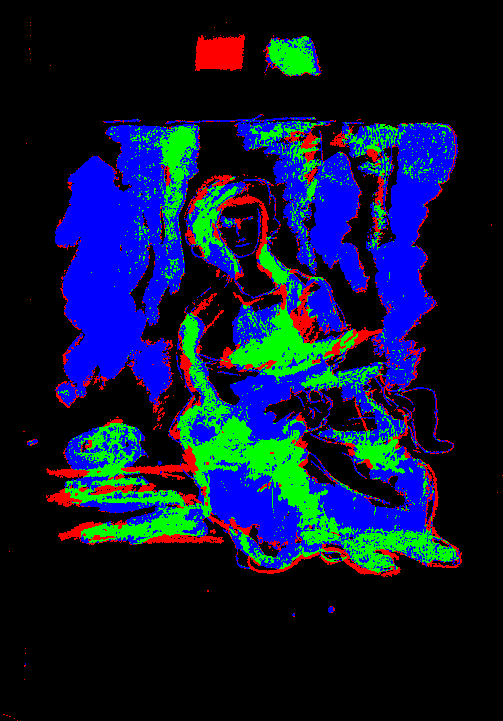}}
		\centerline{(c)}\medskip
	\end{minipage}
	\hspace{1.5cm}
	\begin{minipage}[b]{0.325\linewidth}
		\centering
		\centerline{\includegraphics[width=1\linewidth]{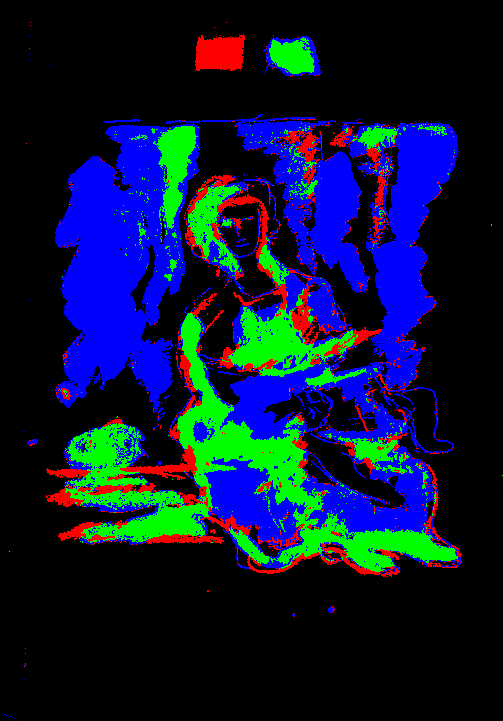}}
		\centerline{(d)}\medskip
	\end{minipage}
	\begin{minipage}[b]{0.325\linewidth}
		\centering
		\centerline{\includegraphics[width=1\linewidth]{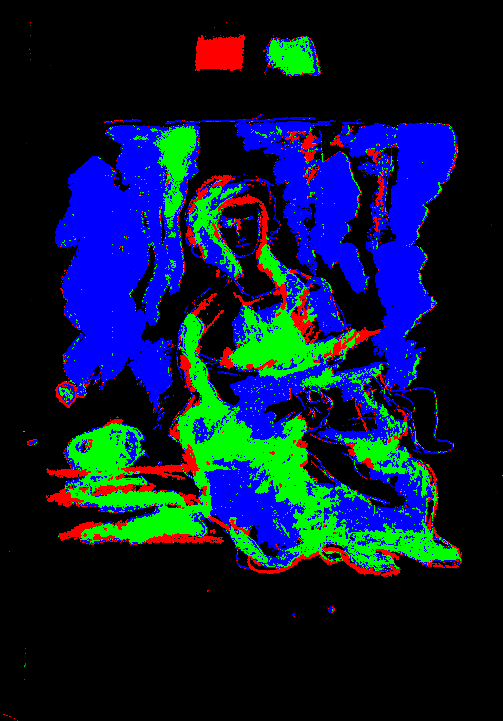}}
		\centerline{(e)}\medskip
	\end{minipage}
	\caption{Label maps. (a) Ground truth, (b) SimRGB, (c) HSI-h, (d) HSI-EMAP, (e) HSIhSI-EMAP.}
	\label{Qualitative}
\end{figure}

The results are qualitatively compared in Fig.~\ref{Qualitative}. In this
figure, (a) represents the ground truth (GT). Red color in GT corresponds to
the red chalk, green represents the red chalk that is overlaid by the black ink
and blue color, is the black ink class. Black color in GT represents the
background and is not considered during the classification. As it can be
observed from this image, SimRGB label map contains high portion of
misclassification, which is highly improved by HSI-h, HSI-EMAP and HSIhSI-EMAP.

\section{Conclusion}

In this work, we propose and evaluate a hyperspectral imaging pipeline for
decomposing the layers of old Master drawings. Our particular focus lies on
distinguishing the commonly used red chalk and black ink.
We propose two descriptors to the field of hyperspectral historical document
analysis, namely hyper-hue and extended multi-attribute profile. We also
address focus shifting, an artifact in hyperspectral imaging, by focus
stacking.

Our comparative results confirm that hyperspectral images are at identical
resolution and SNR more informative than RGB images and result in better layer
separation performance. EMAP and hyper-hue both outperform the raw
hyperspectral features, and focus stacking of hyperspectral images positively
impacts the layer separation.





\bibliographystyle{MyIEEEtran}
\bibliography{ref_DAS}

\end{document}